\documentclass[10pt,twocolumn,letterpaper]{article}

\usepackage{wacv}              % To produce the CAMERA-READY version
\usepackage{graphicx}
\usepackage{amsmath}
\usepackage{amssymb}
\usepackage{booktabs}

\usepackage[pagebackref,breaklinks,colorlinks]{hyperref}

\usepackage{graphicx}
\usepackage{float}
\usepackage[section]{algorithm}
\usepackage{algpseudocode}
\usepackage{pythonhighlight}
\usepackage{amssymb}
\usepackage{amsmath}
\usepackage{multirow}
\usepackage{booktabs}
\usepackage{comment}
\usepackage{amsmath}
\usepackage{oubraces}
\usepackage{colortbl}
\usepackage{enumitem}
\usepackage{xcolor}
\usepackage[accsupp]{axessibility}

\usepackage[capitalize]{cleveref}
\crefname{section}{Sec.}{Secs.}
\Crefname{section}{Section}{Sections}
\Crefname{table}{Table}{Tables}
\crefname{table}{Tab.}{Tabs.}

\def\wacvPaperID{*****} % *** Enter the WACV Paper ID here
\def\confName{WACV}
\def\confYear{2024}

\begin{document}

\title{GIPCOL: Graph-Injected Soft Prompting for Compositional Zero-Shot Learning}

\author{Guangyue Xu$^{1}$, Joyce Chai$^{2}$, Parisa Kordjamshidi$^{1}$\\
$^{1}$ Michigan State University $^{2}$ University of Michigan \\
{\tt\small \{xuguang3, kordjams\}@msu.edu, chaijy@umich.edu} \\
% {\tt\small masi@di.uniroma1.it  \{renzhiy1, groszst, liuxm\}@msu.edu}
}
\maketitle

%%%%%%%%% ABSTRACT
\begin{abstract}
Pre-trained vision-language models (VLMs) have achieved promising success in many fields, especially with prompt learning paradigm.
%\pk{However, designing proper textual prompts to adapt VLMs for downstream tasks is still challenging. : remove}
In this work, we propose \textit{GIPCOL} (\textbf{G}raph-\textbf{I}njected Soft \textbf{P}rompting for \textbf{CO}mpositional \textbf{L}earning) to better explore the compositional zero-shot learning (CZSL) ability of VLMs within the prompt-based learning framework.
The soft prompt in \textit{GIPCOL} is structured and consists of the prefix learnable vectors, attribute label and object label. 
In addition, the attribute and object labels in the soft prompt are designated as nodes in a compositional graph. 
The compositional graph is constructed based on the compositional structure of the objects and attributes extracted from the training data and consequently feeds the updated concept representation into the soft prompt to capture this compositional structure for a better prompting for CZSL.  
With the new prompting strategy, \textit{GIPCOL} achieves state-of-the-art AUC results on all three CZSL benchmarks, including MIT-States, UT-Zappos, and C-GQA datasets in both closed and open settings compared to previous non-CLIP as well as CLIP-based methods. We analyze when and why \textit{GIPCOL} operates well given the CLIP backbone and its training data limitations, and our findings shed light on designing more effective prompts for CZSL.

\end{abstract}

\maketitle

\section{Introduction}

\begin{figure}[t]
 \begin{center} 
  \includegraphics[width=0.5\textwidth]{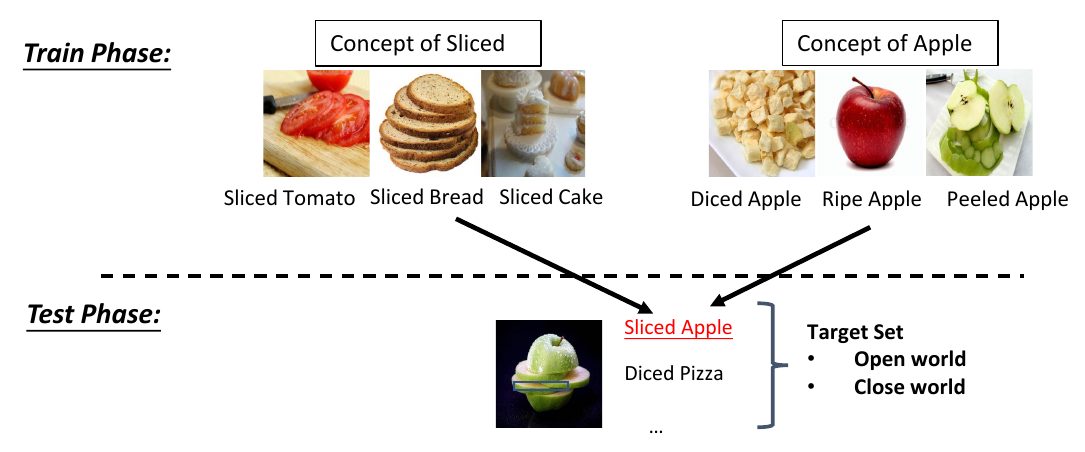}
  \caption{CZSL setting: given the element concepts of \textit{sliced} and \textit{apple}, our target is to recognize the compositional concept \textit{sliced apple}.} %from given target set (close world) or all attr-obj combinations (open world).}
  \label{fig:example}
  \end{center}
\end{figure}

Compositional ability is a key component of human intelligence and should be an important building block for current autonomous AI agents. Fig.~\ref{fig:example} demonstrates a compositional learning example where after learning the element concepts \textit{sliced} and \textit{apple}, the autonomous agent is expected to recognize the novel composition \textit{sliced apple},  by composing the leared element concepts\footnote{element concepts also known as primitive concepts including both attributes and objects in CZSL} which has not been observed during the training time.
This example shows the compositional attribute-object learning problem and this type of compositional ability is essential for language grounding in the vision-language tasks, such as instruction following \cite{chai_lang}, navigation \cite{navigation} , and image captioning~\cite{img_caption}. 

In this paper, we investigate the compositional zero-shot  learning (CZSL) problem as shown in the example. It  requires agents to recognize novel compositions of the attribute-object (attr-obj) pairs appearing in an image by composing previously learned element concepts (e.g., ``sliced'' and ``apple'' individually are considered as element concepts).
The main challenges of CZSL are
1) zero-shot setting in which we do not have training data for the novel compositions.
2) the model should learn the compositional rules to compose the learned element concepts.
3) the distribution shift from the training data to the test data cased by zero-shot setting. Such  shift causes the learned models overfitting the seen compositions and makes it difficult to generalize to  novel compositions.
Previous solutions usually construct a shared embedding space to calculate the matching scores between images and seen pairs and add different generalizing constraints to regularize the space expecting the learnt embeddings capable of encoding compositional properties \cite{attrasopt,graph_comp,open_world_comp}.
Given impressive performance of large VLMs on downstream tasks, in this work, we attempt to solve CZSL from the lens of prompting large VLMs specifically using CLIP \cite{clip} as in \cite{csp}. 
% We propose \textit{GIPCOL} to improve the compositional learning ability of current VLMs. 

Different from traditional zero-shot learning (ZSL) settings where each class is represented by a single text label~\cite{coop,cocoop}, CZSL needs to consider the compositional information among the concepts.
Therefore, the prompt design which can efficiently encode the compositional information is the main challenge for our work.  We expect the designed prompt can re-program CLIP for compositional learning \cite{re-program} and the compositional labels in the prompt should consider the compositonal information.  
Motivated by above expectations, we propose GIPCOL (Graph-Injected Soft Prompting for COmpositional Learning) to design a better prompt to apply VMLs in CZSL.
The core idea of \textit{GIPCOL} is to re-program CLIP for CZSL by 
setting the prefix vectors in the soft prompt as learnable parameters which is different from CSP~\cite{csp}. 
Moreover, \textit{GIPCOL} captures the compositional structure between concepts by constructing a compositional graph from the seen pairs in the training dataset. 
The concepts, both element concept and compositional concept, are acting as nodes in the graph and the compositional graph models the feasible topological combinations between  these concepts.
\textit{GIPCOL} uses a GNN module to update the element label's representations based on their neighbor information in the constructed compositional graph.  
And the updated element embedding is used as class labels int the soft prompt. 
Concretely, the learnable prefix vectors and GNN-updated element concepts consist of the soft prompt for \textit{GIPCOL} and work together to explore CLIP's knowledge for CZSL.  
%the learnable prefix vectors in the soft prompt is used to replace the CLIP's hard prompting words ('a photo of') to increase CLIP's compositional learning capacity.
%{\jycc the following sentence is a repitition from the previous sentence?}
%The GNN module captures the compositional interactions between concepts and updates element concept representations from neighbors in the compositional graph.
%The role of soft-embedding is similar to a verbalizer in the general NLP prompt-learning framework~\cite{liupengfei}. 
%\pk{Here you use the jargon like soft-embedding and soft-prompt but it is not clear what you mean. }
The contributions of this work can be summarized as follows,

\begin{itemize}

    %\item We are among the first works to utilize the structural compositional information to design the soft prompt representation. Although we utilize GNN to extract the compositional information, it is a pluggable module that can be replaced by other differentiable architectures as long as they can model the compositional interactions among element concepts.
    
    \item \textit{Novel prompting design}. Our technique introduces a novel way of utilizing the compositional structure of concepts for constructing the soft prompts. Though we use GNN for capturing this structure, any other differentiable architectures can be used here to enrich the prompt's compositional representation.

    %\item Inherited from prompt-based methods, \textit{GIPCOL} is a \textit{parameter-efficient} learning framework which can improve CZSL using VLMs without the overhead of fine-tuning the entire model.
    
    %\item Different from previous prompting techniques, in order to address the CZSL's challenges, \textit{GIPCOL} introduces two learnable components, soft-prompting consisting of the object and attribute represetnations and a GNN module. The soft-prompting module reprograms VLMs to address the compositional challenge and the GNN module improves the element concept's embedding by considering the concept's neighbor's information in the compositional graph.% COmpositional graph is confusing I think, I am not sure if you defined it in the intro, maybe say:  in a global GNN that connects all concepts and attributes to their possible compositions.}. 
    
    \item \textit{GIPCOL achieves SoTA AUC results on all three CZSL benchmarks}, including MIT-States, UT-Zappos, and the more challenging C-GQA datasets. Moreover, it shows consistent improvements compared to other CLIP-based methods on all benchmarks.

    %\item Besides the new prompting design and SoTA performance, our contributions also include a detailed analysis of prompting-based methods in CZSL. In particular, one important finding is that: injecting additional compositional information could help CLIP recognize compositions that have not been seen during the training time and \textit{GIPCOL} proposes an effective way to add such compositional information
\end{itemize}

%\pk{did you make this parameter efficient? or are you talking bout the whole idea of prompt-based fine tuning? You need to re-phrase your contribution here and clarify.}

%\pk{These claims look superficial and  a very specific terminology is used which is not necessarily clear for the reader to understand the actual contribution.}
%\jycc{Echo Parisa, you need to define/explain these two terms first before talking about their advantages.}

%\pk{is using soft-embedding introduced by you? it is not clear what is your contribution again here, you need to make it clear.}

%\pk{REvised the contribution. Providing analysis can be removed, it is not a standalone contribution, I think: }

\section{Related Work}

\noindent\textbf{Compositional Zero-Shot Learning} (CZSL) is a special field of Zero-Shot Learning (ZSL). 
%\pk{Given a set of attribute-object pairs and related images during the training time, the goal of CZSL is to recognize the novel compositions by learning to compose the learned element attribute/object concepts at inference time.: if you need space this can be removed. its repeated in the intro}
The CZSL is a challenging problem as it requires generalization from seen compositions to novel compositions by learning the compositional rules between element concepts.  There are mainly four lines of research to address this problem.
1) Classifier-based methods train classifiers for attributes and objects separately and combine the element predictions for compositional predictions~\cite{redwine}. 
2) Embedding-based methods construct a shared embedding space for both textual pairs and images. Different methods add different constraints on the space to enhance compositionality~\cite{attrasopt}.
3) Generation-based methods learn to generate visual features for the novel compositions and train classifiers from the generated images~\cite{xian_feat_gen}.
4) Newly proposed prompt-based methods utilize CLIP and introduce learnable element concept embedding or soft prefix vectors in the soft prompt to solve CZSL problems~\cite{csp,prompt_comp}. 
%\pk{make sure you include blank before the citation.}

\noindent\textbf{Prompt-based Learning.}
Parallel to 'fine-tuning', prompt learning provides an efficient mechanism to adapt large pretrained language models(PLMs) or vision-language models (VLMs) to downstream tasks by treating the input prompt as learnable parameters while freezing the rest of the foundation model. Prompt learning is a parameter-efficient framework originated from the NLP field  aiming at utilizing knowledge encoded in PLMs for downstream tasks ~\cite{liupengfei_prompt_survey,gpt,t5}.
% Prompt-based learning usually includes prompt engineering and verbalizer.   
%Prompt-engineering uses a template to transform the original input into a textual prompt
%which usually has some unfilled slots. 
%Then the large frozen language model is used to probabilistically fill or generate the slots. 
% Verbalizer participates later to map the slot to the final output. 
Recently, as the prevalence of large  vision-language models (VLMs),  prompt learning is introduced into multimodal settings to solve VL-related problems~\cite{frozen,pica,million_prompt}, including the CZSL problems~\cite{csp,prompt_comp}.
In both linguistic and multi-modal settings, prompt engineering plays an important role. How to design a suitable prompt template for downstream tasks is a challenge and  \textit{GIPCOL} proposes a novel approach to address this challenge.

\noindent\textbf{Vision-Language Models.} 
Large VMLs are pre-trained to learn the semantic alignment between vision and language modalities in different levels \cite{align,clip}.
Attention-based encoder, large mini-batch contrastive loss, and web-scaled training data are the main factors to boost the performance of such vision-language models.
Recent advances in these pre-trained VLMs have presented a promising direction to promote open-world visual understanding with the help of language. Besides the open-world image classification, VLMs are used in other visual fields, like dense prediction~\cite{denseclip} and caption generation~\cite{clipcap}.

Among existing methods, the most relevant to ours are CSP~\cite{csp} and CGE\cite{graph_comp}.
CSP treat the element concept labels as learnable parameters to prompt CLIP for CZSL and can be considered as a baseline for \textit{GIPCOL}.
CGE encodes compositional concepts using GNN and constructs a shared embedding space to align images and compositional concepts. It is a task-specific architecture and needs to fine-tune the visual encoder to achieve satisfactory performance. 
Compared with such task-specific models, \textit{GIPCOL} is a general prompting method and uses GNN to capture interactions among the concepts for its soft prompting design.  \textit{GIPCOL} fixes CLIP's pre-trained visual and textual encoders and achieves better performance in a more general and parameter-efficient manner. 
It is worth noting that GNN used in CGE and \textit{GIPCOL} have different nature, CGE for compositional encoding and \textit{GIPCOL} for soft prompt construction. 

\begin{figure}[!htb]
 \begin{center} 
  \includegraphics[width=0.75\linewidth]{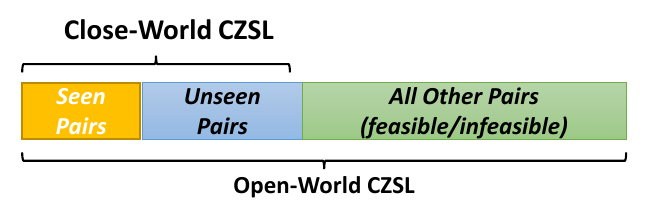}
  \caption{Illustration of different CZSL settings  based on the target compositional set. \textit{GIPCOL} is evaluated under closed-world and open-world settings.}
  \label{fig:czsl_setting}
  \end{center}
\end{figure}

\begin{figure*}[ht]
\includegraphics[width=\textwidth]{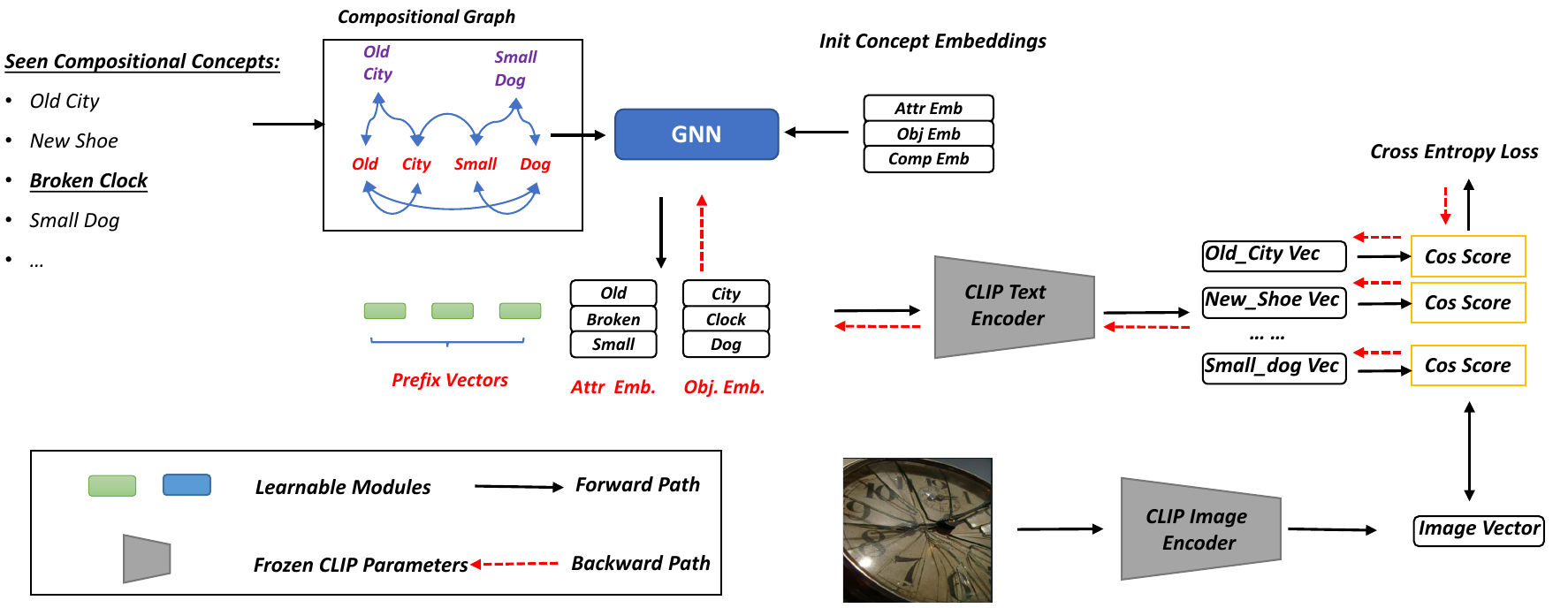}
\centering
\caption{\textit{GIPCOL} Architecture. Besides CLIP's frozen text and visual encoders, \textit{GIPCOL} consists of two learnable components: a soft-prompting module and a GNN. \textit{GIPCOL} calculates the cosine similarity between the given image and all candidate pairs and the cross-entropy loss is
back-propagated through the frozen LM in order to update soft-prompt and GNN.}
\label{fig:arch}
\end{figure*}

%\pk{You need to point to the related work on zappos and MIT datasets somewhere, I did nto see it in intro nor in this section.}

\section{Problem Formulation\label{problem_setting}}
In this section, we formally define the CZSL task.
Let $\mathbb{A}=\left\{a_{0}, a_{1}, \ldots, a_{n}\right\}$ be the attribute set and $\mathbb{O}=\left\{o_{0}, o_{1}, \ldots, o_{m}\right\}$ be the object set. 
All possible compositional label space $\mathbb{C}$ is the Cartesian product of these two element concept sets, $\mathbb{C}=\mathbb{A} \times \mathbb{O}$ with size $n \times m$.
%within which some combinations are not feasible. Following \cite{graph_comp,csp}'s work, we also introduce the feasibility score to filter out infeasible compositions during test time.
At training time, we are given a set of seen\footnote{seen examples also mean training examples, we use them interchangeably in this work.} examples  $\mathbb{C}_{\text {seen }}=\left\{\left(x_{1}, c_{1}\right), \ldots,\left(x_{k}, c_{k}\right)\right\}$, where $x_i$ is an image and $c_i=(a_i,o_i)$\footnote{We use the pair index to denote the object and attribute indexes for the sake of simple notation. The object and attribute indexes do not refer to their original sets in this case. 
} is its compositional label from the seen set $\mathbb{C}_{seen}\subset\mathbb{C}$. 
%is the seen set and a subset from the compositional space $\mathbb{Y}$. 
The goal of CZSL is to learn a function $f$ to assign a compositional label from the target set $\mathbb{C}_{target} \subseteq \mathbb{C}$ to a given image . 
Based on different target set settings as shown in  Fig.~\ref{fig:czsl_setting}, CZSL can be categorized into 
%1) Standard CZSL, where $\mathbb{C}_{target} = \mathbb{C}_{unseen}$ and $\mathbb{C}_{seen} \cap \mathbb{C}_{unseen}=\varnothing$ where the target set only consists of  feasible unseen  pairs introduced in \cite{attrasopt}; 
1) \underline{Closed-world CZSL}, where $\mathbb{C}_{target} = \mathbb{C}_{seen} \cup \mathbb{C}_{unseen}$, the target set consists of both seen and unseen  pairs as introduced in \cite{modu_net}. In this setting, both seen and unseen pairs are feasible. This setting is called a closed-world setting because the test pairs are given in advance.
2) \underline{Open-world CZSL}, where $\mathbb{C}_{target} = \mathbb{C}$. The target set contains all attr-obj combinations including both feasible and infeasible pairs. This is the most challenging case introduced in \cite{open_world_comp}. We evaluate our models  under both closed-world and open-world settings.

\section{GIPCOL}

By pre-training on $400$ million image-text association pairs, \textit{CLIP} has already learned the general knowledge for images recognition. 
In order to fully utilize \textit{CLIP}'s capability in compositional learning, \textit{GIPCOL} freezes CLIP's textual and visual encoders and focuses on structuring its textual prompt to address compositional concept learning. The \textit{GIPCOL}'s architecture is shown in Fig.~\ref{fig:arch}.
In particular, \textit{GIPCOL} adds two learnable components to construct the soft prompt for CZSL: the learnable prefix vectors
%where $k$ is the number of prefix tokens 
and the GNN module.
The prefix vectors are used to add more learnable parameters to represent the compositional concepts and reprogram CLIP for compositional learning.  
% \pk{Different from previous usage~\cite{graph_comp,graph_emb_open_pami}, GNN here plays the role of automating prompt engineering and its output is part of soft prompt which goes through CLIP for compositional learning. Based on the data processing flow, we will describe the prefix vectors, GNN encoder, and CLIP's visual and textual encoder. :
The GNN module is to capture the compositional structure of the objects and attributes for a better compositional concept representation in the constructed soft prompt.
% Our new idea is to use the output of GNN and feed it to the  soft prompt of CLIP to capture the compositional structure in prompt learning. 
We describe the details of \textit{GIPCOL}, including the learnable prefix vectors, GNN, and CLIP's visual/textual encoder in the following section.

\subsection{\textbf{GIPCOL} Architecture\label{sec:arc}}
\vspace{1pt}
\noindent\textbf{Learnable Prefix Vectors.} 
We designate $k$ learnable prefix vectors $\Theta= \{\theta_1, \theta_2, ..., \theta_m\}$ where $\theta_i \in \mathbb{R}^d$ in soft prompt for compositional concept encoding.
$d$ is set to $768$ to be consistent with CLIP embedding size. 
Here, larger $k$ means more learnable parameters and learning ability for compositional concept representation. 
These vectors are used to prepend to the attr-obj embeddings and act as part of  the compositional representation.  
These prefix vectors are fine-tuned by gradients flowing back through CLIP during the training time.

\vspace{1pt}
\noindent\textbf{GNN as Concept Encoder}.
Different from traditional zero-shot learning (ZSL) problems where output labels are treated independently, CZSL requires modeling the interactions between element concepts. For example, given the compositional concept \textit{red apple}, we need to learn both the concept \textit{apple} and how \textit{red} changes \textit{apple}'s state instead of treating red and apple as two independent concepts.
Graph Neural Networks (GNN) have been proved to be able to capture such 
dependencies~\cite{graph_comp,graph_emb_open_pami}.
We introduce GNN in \textit{GIPCOL} to enrich the concept's representations by fusing information from their compositional neighbors as follows,

\begin{equation}
(\hat{a_i},\hat{o_i}) = GNN_{\Phi}(a_i, o_i)
\label{eq:gnn}
\end{equation}

\noindent where $\Phi$ is GNN's parameter, $(a_i, o_i)$ and $(\hat{a_i}, \hat{o_i})$ are the original and updated compositional concept's representation. The updated node representations from GNN will serve as class labels in soft prompt. The whole soft prompt represents the compositional concept and will be put into  CLIP's textual encoder for compositional learning.

\vspace{1pt}
\noindent\textbf{Frozen CLIP's Text Encoder.} 
After obtaining the updated compositional representations $(\hat{a_i},\hat{o_i})$, \textit{GIPCOL} adds the learnable prefix vectors $\Theta = [\theta_1, \theta_2, ..., \theta_m]$ prepending in front of $(\hat{a_i},\hat{o_i})$ to represent compositional concept as follows,

\begin{equation}
\setlength\abovedisplayskip{0pt}
[\underbrace{SOS, \hspace{10pt} \overbrace{\theta_1, \theta_2, ...,\theta_m}^{\textbf{prefix Vectors}}, \hspace{4pt} \overbrace{\hat{a_i}, \hat{o_i},}^{\textbf{GNN-Updated Concept}} EOS}_{\textbf{Soft Prompt \noindent as Compositional Concept Representation}}].
\label{eq:prompt}
\end{equation}

\noindent Then we use CLIP's frozen text encoder, a Bert encoder~\cite{bert}, to extract the normalized \textit{EOS} vector as the compositional concept's representation  for further multi-modal alignment as follows,

\begin{equation}
\setlength\abovedisplayskip{0pt}
%\quad 
\boldsymbol{c}_{i}=\frac{TxtEnc\left({\Theta, (\hat{a_i},\hat{o_i})}\right)}{\left\|TxtEnc\left(\Theta,(\hat{a_i}, \hat{o_i})\right) \right\|}
\label{eq:txt_encoder}
\end{equation}

\noindent where $(\hat{a_i},\hat{o_i})$ is the GNN-updated attribute and object vectors and $c_i$ is the $i$-th compositional concept vector encoded by CLIP.

\vspace{1pt}
\noindent\textbf{Frozen visual encoder.}  
Following CLIP's pre-processing routine, we first rescale the image's size to $224 \times 224$. Then we use \textit{ViT-L/14} as the visual encoder ViT to encode the image and extract the \textit{[CLASS]} token as the image's representation. The extracted image vector $x_i$ needs to be normalized as follows for further similarity calculation. 
%\pk{double check if you have described/cited this before}
\begin{equation}
\setlength\abovedisplayskip{0pt}
\boldsymbol{x}_{i}=\frac{VisEnc(v_i)}{\left\|VisEnc(v_i)\right\|}
\label{eq:vis_encoder}
\end{equation}

\noindent where $v_i$ is the given image and $x_i$ is its vector representation.

\vspace{1pt}
\noindent\textbf{Aligning Image and Compositional Concept.} 
After obtaining the vectors for the compositional concept $c_i$ and the image $x$, \textit{GIPCOL} calcualtes the probability of $x$ belonging to class $c_i$ as follows, 

\begin{equation}
\setlength\abovedisplayskip{0pt}
p(c_i \mid x)=\frac{\exp \left(\left(x \cdot c_i\right) / \tau\right)}{\sum_{k=1}^{K} \exp \left(\left(x \cdot c_k\right) / \tau\right).}
\label{eq:sim}
\end{equation}

\noindent where $\tau$ is a temperature parameter from CLIP, $\cdot$ denotes the inner product of the concept vector and the image vector and $K$ is the number of attr-obj pairs in the training set.

\subsection{GNN in Soft Prompting\label{gnn}}

As disussed previously, a key idea to address the CZSL problem is to learn concept representations that are able to internalize the compositional information.  
Graph could be the tool to model such compositional dependencies.
And this idea has been used in previous work~\cite{graph_comp,graph_emb_open_pami} by applying Graph Neural Networks(GNN) as encoders to represent the compositional concepts. 
Although we adopt similar graph-based methods for compositional encoding, our novelty is to use the graph's compositional structure to facilitate the automated prompt engineering in compositional learning as shown in Appendix \ref{gnn_compare}.
We model the  element concepts and their composition explicitly in GNN for the soft prompting construction.
In principle, the GNN module can be replaced by other differentiable architectures that are able to capture the concept's compositional information. 
%Concretely, a compositional graph is represented by $\mathcal{G}=(\mathcal{V}, \mathcal{E})$ where $\mathcal{V}=\left\{v_1, \ldots, v_N\right\}$ is the set of concept nodes and $\mathcal{E}=\left\{e_1, \ldots, e_M\right\}$ is the set of edges representing compostional relation.
We describe the detailed GNN application in \textit{GIPCOL} next.

\noindent\textbf{Node Embedding $\mathcal{V}$.\label{com_graph}} There are two types of nodes in \textit{GIPCOL}'s compositional graph: element concept node and compositional concept node. The node embedding's size is $R^{(|a|+|o|+|c|)*d}$, where $|a|$ is the attribute number, $|o|$ is the object number, $|c|$ is the training pair number and $d$ is the feature dimension.
For the element nodes, we initialize them using CLIP's embedding vectors. 
For the compositional nodes, we initialize them using the average embedding of the element nodes, that is, $\frac{att\_vec + obj\_vec}{2}$. 
\textit{GIPCOL} relies on GNN to fuse information from the constructed compositional graph and update the concept's representation.

%If the length of the tokenized element concept is bigger than 1. We use the average embedding as the node representation. \jycc{the above two sentences not clear, if itself cannot be a sentence.}During training time, we 
%update the element embedding based on compositional graph information. 
%And this improve the model's compositional ability. 

\begin{figure*}[!htb]
\small
\begin{center} 
\includegraphics[width=\linewidth]{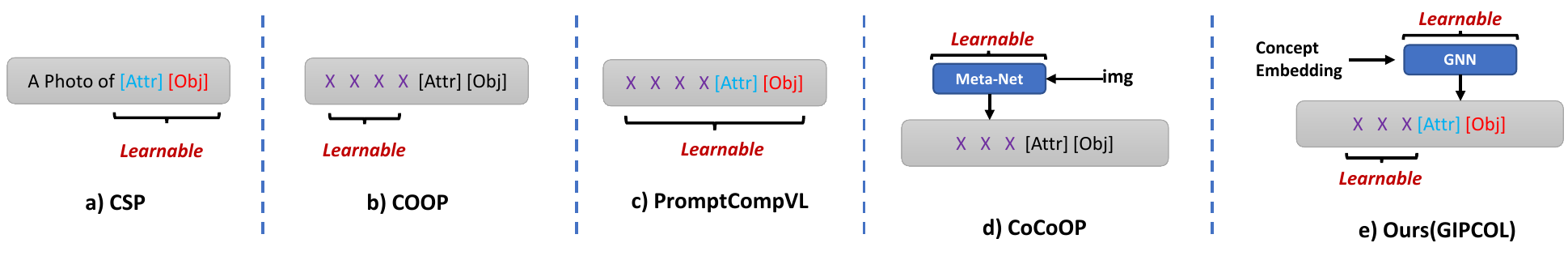}
\end{center}
\caption{Different prompting strategies. \textit{GIPCOL} combines both soft prefix vector and GNN for prompt construction.}
\label{promptType}
\end{figure*}

\noindent\textbf{Compositional Graph Constructions $\mathcal{E}$.} We use a graph to capture the compositional dependencies and learn richer concept representations. 
The connection design among concepts is the key challenge for such graph.
In order to utilize the feasible compositional information, \textit{GIPCOL} considers the training pairs and construct one single compositional graph for both closed-world CZSL and open-world CZSL to conserve the computing and storage resources. Specifically, given a pair $c=(a,o)$, besides the self-connected edge, \textit{GIPCOL} adds three undirected edges $(c\leftrightarrow a)$, $(c \leftrightarrow o)$ and $(a\leftrightarrow o)$ in the graph where the adjacency matrix $A \in \mathbb{R}^{K \times K}$  is symmetric with $K=|a|+|o|+|c|$.
The compositional concept plays the bridging role to help connect element concepts and only the element concepts are used to construct the compositional prompting due to the zero-shot setting.

\noindent\textbf{GNN Module:}  Once we have the compositional graph and the initialized concept features, we can update the concept's embedding by fusing the compositional information from its neighbors. Any GNN models could be applied here and in \textit{GIPCOL}, we use Graph Convolution Network (GCN)~\cite{gcn} in Eq.~\ref{gcn} for compositional encoding.

\begin{equation}
H^{(l+1)}=\sigma\left(\tilde{D}^{-\frac{1}{2}} \tilde{A} \tilde{D}^{-\frac{1}{2}} H^{(l)} \Phi^{(l)}\right)
\label{gcn}
\end{equation}

\noindent where 
$H^l$ denotes the node's representations in the $l^{th}$ layer, 
$\sigma$ is the non-linearity ReLU function, 
$\tilde{A}$ is the adjacency matrix with added self-connections,
$\tilde{D}$ is a diagonal node degree matrix 
and $\Phi^l$ is the learnable weight matrix in layer $l$.
Notably, other graph constructing methods, like using external knowledge \cite{kg-sp}, and other GNN models, like GAT \cite{gat}, could be further explored to improve CZSL performance based on \textit{GIPCOL}'s architecture. However, these are not target of this work. Here, \textit{GIPCOL} shows the effectiveness of utilizing compositional knowledge in prompting construction in CZSL.

\subsection{Training}
After obtaining the concept and image representations, we calculate the class probability using Eq.~\ref{eq:sim}. 
And the regularized Cross-Entropy loss is used to update \textit{GIPCOL}'s prefix vectors $\Theta$ and GNN parameters $\Phi$ as follows,

\begin{equation}
-\frac{1}{\left|\mathbb{N}\right|} \sum_{i \in \mathbb{N}} \log p_{\boldsymbol{\theta}}(c_i \mid x)+\lambda_1\|\boldsymbol{\Theta}\|^2 +\lambda_2\|\boldsymbol{\Phi}\|^2
\label{cross_entropy}
\end{equation}

\noindent where $\lambda_1$ and $\lambda_2$ are the hyper-parameters to control the weight decay for prefix vector and GCN separately.
\textit{GIPCOL} keeps CLIP's pre-trained textural and visual encoders fixed during the training time. And more details about the  training process can be found in Append.~\ref{alg_appendix}.

\subsection{Inference\label{infer}}
%As discussed in Sec.\ref{problem_setting}, we test \textit{GIPCOL} under both closed-world and open-world settings. 
During inference, given an image,  we first construct the soft prompts for all target concepts using the fine-tuned prefix vectors and GNN. Then, we use CLIP's frozen textual and visual encoders to obtain the image vector $x$ and the target concept vector set $\mathbb{C}_{tareget}$. Then we use cosine measurement to select the most similar attr-obj pair from $\mathbb{C}_{target}$ as the compositional label as follows,

\begin{equation}
\setlength\abovedisplayskip{0pt}
\hat{c}=\underset{c_i \in \mathbb{C}_{\text {target}}}{\arg \max }\ cos \left(c_i,x\right).
\label{eq:infer}
\end{equation}
\noindent where $c_i$ is the $i$-th compositional vector from the target set.

\subsection{CLIP-Prompting Method Comparison}

In this section, we clarify the difference between all CLIP-prompting methods used in CZSL as shown in Fig.~\ref{promptType}. 
Generally, all current CLIP-prompting methods keeps the image representation fixed and learn constructing the CLIP's textual prompt to represent the compositional concept as shown in Eq.~\ref{eq:prompt}. 
The main difference is that CSP\cite{csp} learns the element embedding, COOP\cite{coop} learns the prefix vectors and PromptCompVL learns both the element embedding and the prefix vectors. 
All these three methods do not explicit consider the compositional structures between concepts. 
In order to inject more semantic information into soft prompt, CoCoOP\cite{cocoop} introduces a Meta-Net and tries to modify the prefix vectors based on each image input. 
It uses the instance-level information not the global compositional information for CZSL. Such instance-level prompting also causes training inefficient and consumes a significant amount of computing resources as discussed in that work. 
Different from all previous methods, \textit{GIPCOL} proposed a novel prompting strategy by combining the learnable prefix vectors and the GNN module and the detailed comparition is in Append. \ref{gnn_compare}.
%\pk{This looks like a related work, why suddenly appear here? NEEDS to be moved to somewhere relevant!} 
% In order to explore CLIP's compositional learning ability, all the methods use different ways to construct the text input for CLIP. 

\section{Experiments}

\subsection{Experimental Setting}

%We use two datasets to check \textit{GIPCOL}'s CZSL ability: 
\noindent\textbf{Datasets.} We conduct experiments on three datasets, MIT-States~\cite{mitstates} UT-Zappos~\cite{zappos} and C-GQA \cite{graph_comp}. 
MIT-States and C-GQA consist of images with objects and their attributes in the general domain. In contrast, UT-Zappos contains images of shoes paired with their material attributes which is a more domain-specific dataset.
Our experiments follow the previous works \cite{modu_net,graph_comp} on the data split for training and testing. More details about the data splits and statistics can be found in Append.~\ref{data_stat_append}.

\vspace{3pt}
\noindent\textbf{Implementation details.}
We extend on the codebase of  \cite{csp}\footnote{https://github.com/BatsResearch/csp} and ~\cite{graph_comp}\footnote{https://github.com/ExplainableML/czsl} for \textit{GIPCOL}'s implementation.
Moreover, for a fair comparison, %the context length is set to 
the length of the prefix vector, $k$, is set to $3$ which is the same length of  CLIP hard-prompting \textit{'a photo of'}. The dimension of soft-prompting $d$ is set to $768$ which is consistent with CLIP's model setting. Moreover, we use two-layer GCN to encode concepts and the corresponding  GNN's learnable parameters are $\Phi = \{\Phi^1, \Phi^2\}$ 
%In the open setting,  we introduce the feasibility score to adjust the unseen pairs' logits as discussed in the previous section.
Our code will be made publicly available on GitHub\footnote{\url{https://github.com/HLR/GIPCOL}}.  
%{\color{red}{please make the links clickable through the paper. Also, the github is again empty. Now that you have time, please finalize the github codes for both papers ASAP, please DO NOT postpone this.}}

\noindent\textbf{Evaluation Metrics}.
Zero-shot models are biased to the seen classes as shown in previous woks \cite{gzsl,open_world_comp}. 
As a standard method in zero-shot learning, we introduce a scalar value adding to the unseen classes to adjust the bias towards the seen classes as used in \cite{modu_net,csp}. 
By varying the added bias from $-\infty$ to $+\infty$, we report \textit{GIPCOL}'s performance using the following four metrics in both the closed-world and the open-world settings as discussed in Sec.~\ref{problem_setting}:
1) \textit{Best seen accuracy} (S), testing only on seen compositions when bias is  $-\infty$;
2) \textit{Best unseen accuracy} (U), testing only on unseen compositions when bias is $+\infty$;
3) \textit{Best harmonic mean} (HM) which balances the performance between seen and unseen accuracies;
4) \textit{Area Under the Curve} (AUC), the area below the seen-unseen accuracy curve by varying the  scalar added to the unseen compositional concepts.

\noindent\textbf{Baselines.}
We compare \textit{GIPCOL} with two types of baselines: 
1) \textit{non-CLIP methods} (top seven models in the closed setting and top six in the open setting) namely Attributes as Operators
(AoP)\cite{attrasopt}, Label Embed+ (LE+)\cite{redwine}, Task Modular Networks
(TMN)\cite{modu_net}, SymNet\cite{sym_net}, Compositional Graph
Embeddings (CGE)\cite{graph_comp}, Compositional Cosine Logits
(CompCos)\cite{open_world_comp} and Siamese Contrastive Embedding Network(SCEN)\cite{scen}.
2) \textit{CLIP-based methods} (the bottom three models), namely CLIP\cite{clip}, Context Optimization(COOP)\cite{coop} and compositional soft prompting (CSP)\cite{csp}. 
%The results are shown in Table~\ref{tab:close_result} and Table~\ref{tab:open_result}. 

\begin{table*}[!htbp]
\centering
\small
\resizebox{0.8\textwidth}{!}{
\begin{tabular}{ccccccccccccc} 
\hline
& \multicolumn{4}{c}{\textbf{MIT-States}} & \multicolumn{4}{c}{\textbf{UT\_Zappos}} & \multicolumn{4}{c}{\textbf{C-GQA}} \\
\cmidrule(r){2-5}\cmidrule(r){6-9}\cmidrule(r){10-13}
Method & S & U & H & AUC & S & U & H & AUC & S & U & H & AUC \\
\hline
\textbf{AoP}~\cite{attrasopt} & 14.3 & 17.4 & 9.9 & 1.6 & 59.8 & 54.2 & 40.8 & 25.9 & 17.0 & 5.6 & 5.9 & 0.7\\
\textbf{LE+}~\cite{redwine} & 15.0 & 20.1 & 10.7 & 2.0 & 53.0 & 61.9 & 41.0 & 25.7 & 18.1 & 5.6 & 6.1 & 0.8\\ 
\textbf{TMN}~\cite{modu_net} & 20.2 & 20.1 & 13.0 & 2.9 & 58.7 & 60.0 & 45.0 & 29.3 & 23.1 & 6.5 & 7.5 & 1.1\\
\textbf{SymNet}~\cite{sym_net} & 24.2 & 25.2 & 16.1 & 3.0 & 49.8 & 57.4 & 40.4 & 23.4 & 26.8 & 10.3 & 11.0 & 2.1\\
\textbf{CompCos}~\cite{open_world_comp} & 25.3 & 24.6 & 16.4 & 4.5 & 59.8 & 62.5 & 43.1 & 28.7 & 28.1 & 11.2 & 12.4 & 2.6\\
\textbf{CGE}~\cite{graph_comp} & 32.8 & 28.0 & 21.4 & 6.5 & 64.5 & \cellcolor{lightgray}71.5 & \cellcolor{lightgray}60.5 & 33.5 & 33.5 & 15.5 & 16.0 & 4.2\\
\textbf{SCEN}~\cite{scen} & 29.9 & 25.2 & 18.4 & 5.3 & 63.5 & 63.1 & 47.8 & 32.0 & 28.9 & 25.4 & 17.5 & 5.5\\
%\textbf{CANet}(23) & 29 & 26.2 & 17.9 & 5.4 & 61.0 & 66.3 & 47.3 & 33.1 & 30.0 & 13.2 & 14.5 & 3.3\\
%\textbf{ADE}(23) &  &  &  &  & 63.0 & 64.3 & 51.1 & 35.1 & 35.0 &  & 17.7 & 5.2\\
\hline
\textbf{CLIP}~\cite{clip} & 30.2 & 40.0 & 26.1 & 11.0 & 15.8 & 49.1 & 15.6 & 5.0 &7.5 & 25.0 & 8.6 & 1.4 \\
\textbf{COOP}~\cite{coop} & 34.4 & 47.6 & 29.8 & 13.5 & 52.1 & 49.3 & 34.6 & 18.8 &20.5 & 26.8 & 17.1 & 4.4 \\
\textbf{CSP}~\cite{csp} & 46.6 & \cellcolor{lightgray}49.9 & 36.3 & 19.4 & 64.2 & 66.2 & 46.6 & 33.0 & 28.8 & 26.8 & 20.5 & 6.2 \\
%\textbf{GIPCOL(ours)} & \textbf{48.5} & \textbf{49.6} & \textbf{36.6} & \textbf{19.9} & 64.4 & 64.0 & 46.12 & 32.2 & 31.61 & \textbf{28.41} & \textbf{22.91} & \textbf{7.41}\\
\hline
\textbf{\textit{GIPCOL} (Ours)} & \cellcolor{lightgray}48.5 & 49.6 & \cellcolor{lightgray}36.6 & \cellcolor{lightgray}19.9 & \cellcolor{lightgray}65.0 & 68.5 & 48.8 & \cellcolor{lightgray}36.2 & \cellcolor{lightgray}31.92 &\cellcolor{lightgray} 28.4 & \cellcolor{lightgray}22.5 & \cellcolor{lightgray}7.14\\
\hline
\end{tabular}}
\caption{Closed-World CZSL results on UT-Zappos, Mit-States and C-GQA datasets.}% We report best seen accuracy S, best unseen accuracy U, best harmonic mean(HM) and area under the curve(AUC) for comparison.} %\jycc{within the closed world setting, you have two settings: standard and generalized. which one are these results from? }}
\label{tab:close_result}
\end{table*}

\begin{table*}[!htbp]
%\small
\centering
\resizebox{0.8\textwidth}{!}{
\begin{tabular}{ccccccccccccc} 
\hline
& \multicolumn{4}{c}{\textbf{MIT-States}} & \multicolumn{4}{c}{\textbf{UT\_Zappos}} & \multicolumn{4}{c}{\textbf{C-GQA}} \\
\cmidrule(r){2-5}\cmidrule(r){6-9}\cmidrule(r){10-13}
Method & S & U & H & AUC & S & U & H & AUC & S & U & H & AUC \\
\hline
\textbf{AoP}~\cite{attrasopt} & 16.6 & 5.7 & 4.7 & 0.7 & 50.9 & 34.2 & 29.4 & 13.7 & - & - & - & -\\
\textbf{LE+}~\cite{redwine} & 14.2 & 2.5 & 2.7 & 0.3 & 60.4 & 36.5 & 30.5 & 16.3 & 19.2 & 0.7 & 1.0 & 0.08\\ 
\textbf{TMN}~\cite{modu_net} & 12.6 & 0.9 & 1.2 & 0.1 & 55.9 & 18.1 & 21.7 & 8.4 & - & - & - & -\\
\textbf{SymNet}~\cite{sym_net} & 21.4 & 7.0 & 5.8 & 0.8 & 53.3 & 44.6 & 34.5 & 18.5 & 26.7 & 2.2 & 3.3 & 0.43\\
\textbf{CompCos}~\cite{open_world_comp} & 25.4 & 10.0 & 8.9 & 1.6 & 59.3 & 46.8 & 36.9 & 21.3 & - & - & - & - \\
\textbf{CGE}~\cite{graph_comp} & 32.4 & 5.1 & 6.0 & 1.0 & 61.7 & \cellcolor{lightgray}47.7 & 39.0 & 23.1 & \cellcolor{lightgray}32.1 & 1.8 & 2.9 & 0.47\\
%\textbf{ADE}(23)~\cite{graph_comp} &  &  &  &  & 62.4 & \cellcolor{lightgray}50.7 & 44.8 & 27.1 & \cellcolor{lightgray}35.1 & 4.8 & 7.6 & 1.42\\
%\textbf{KG-SP}~\cite{kg-sp} & 28.4 & 7.5 & 7.4 & 1.3 & 61.8 & \cellcolor{lightgray}52.1 & \cellcolor{lightgray}42.3 & \cellcolor{lightgray}26.5 & 31.5 & 2.9 & 4.7 & 0.78\\
\hline
\textbf{CLIP}~\cite{clip} & 30.1 & 14.3 & 12.8 & 3.0 & 15.7 & 20.6 & 11.2 & 2.2 & 7.5 & 4.6 & 4.0 & 0.27\\
\textbf{COOP}~\cite{coop} & 34.6 & 9.3 & 12.3 & 2.8 & 52.1 & 31.5 & 28.9 & 13.2 & 21.0 & 4.6 & 5.5 & 0.70 \\
\textbf{CSP}~\cite{csp} & 46.3 & 15.7 & 17.4 & 5.7 & 64.1 & 44.1 & 38.9 & 22.7 & 28.7 & 5.2 & 6.9 & 1.20 \\
\hline
\textbf{\textit{GIPCOL} (Ours)} & \cellcolor{lightgray}48.5 & \cellcolor{lightgray}16.0 & \cellcolor{lightgray}17.9 & \cellcolor{lightgray}6.3 & \cellcolor{lightgray}65.0 & 45.0 & \cellcolor{lightgray}40.1 & \cellcolor{lightgray}23.5 & 31.6 & \cellcolor{lightgray}5.5 & \cellcolor{lightgray}7.3 & \cellcolor{lightgray}1.30\\
\hline
\end{tabular}}
\caption{Open-World CZSL results on UT-Zappos, Mit-States and C-GQA datasets.}
\label{tab:open_result}
\end{table*}

\vspace{3pt}
\noindent\textbf{Feasibility Calibration in Open-World Setting.}
Open-world CZSL is more challenging compared with the closed-world setting as the class space contains all possible combinations of attributes and objects including both feasible compositions and infeasible compositions. In order to filter out the infeasible compositions, we apply the feasibility calibration as used in \cite{open_world_comp,csp}.
% Feasibility scores fro all seen pairs are set to $1$.
For each unseen pair $(a,o)$, we first collect two sets from the training data. One is the applicable attribute set $A=\{a_1, a_2, \dots, a_M\}$ for the target object $o$ and the other is the applicable object set $O=\{o_1, o_2, \dots, o_N\}$ for the target attribute $a$ where $(a_i, o)$ and $(a, o_j)$ has been observed in training time. Then we calculate the similarity between $a$ and each element in $A$ and use the maximum similarity score as this pair's  attribute feasibility score as follows,
\begin{equation}
f_a(a, o)=\max_{(a_i, o) \in \mathbb{C}_{\text {seen }}} \frac{e(a) \cdot e(a_i)}{\|e(a)\|\|e(a_i)\|},
\label{attr_fea}
\end{equation}

\noindent where $e$ is the GloVe embedding~\cite{glove}. On the other hand, this pair's object feasibility score is calculated in a similar way based on the applicable object set. 
Finally, the unseen pair feasibility score is calculated as the average of the two scores, $\frac{f_a + f_o}{2}$.
%The feasibility score is calculated in a simple and intuitive manner and its basic assumption is that similar objects share similar states. 
After obtaining the feasibility score for all unseen pairs, we can filter out infeasible compositions by setting a threshold $T$ whcih can be tuned based on the validation set. The final prediction for image $x$ in the open-world setting is computed as follows,

\begin{equation}
\setlength\abovedisplayskip{0pt}
\hat{c}=\underset{c_i \in \mathbb{C}_{\text {target}}, \hspace{2pt} c_i \ge T}{\arg \max }\ cos \left(c_i,x\right).
\label{eq:open_infer}
\end{equation}

\noindent Different from the closed-world setting, we require the feasibility score of the predicted label $c$ to be larger than a threshold. The threshold uses in our experiments is shown in Append.~\ref{feas_score_append}.

\subsection{Results}

\noindent\textbf{Results on MIT-States}. 
As shown in Tab.~\ref{tab:close_result} and \ref{tab:open_result},  %shows the overall results and we can see that
\textit{GIPCOL} achieves the new SoTA results on 
MIT-States on both closed-world and open-world settings compared with CLIP and non-CLIP baselines (except for the best-unseen metric (U)).
The CLIP-based models have consistently better performance compared to the non-CLIP methods\footnote{In principle CLIP-based and non-CLIP-based methods cannot be directly compared as we have no information about the training data used for CLIP training. Here we follow previous work and include these baselines for the sake of comparison and consistency with the previous work.}. 
CLIP-prompting methods, including COOP, CSP and ours, further boost the performance compared to the vanilla CLIP model. 
%This result demonstrates the importance of prompt learning for CZSL.

% \jycc{The following about UT-Zappos should go to a subsection for UT-Zappos}

\vspace{3pt}
\noindent\textbf{Results on UT-Zappos}. On UT-Zappos, previous CLIP-based approaches under-perform the SoTA performance achieved by CGE which is a non-CLIP model. However, \textit{GIPCOL} successfully surpasses the CGE model. Note that UT-Zappos is a domain-specific dataset that consists of shoe types and the materials. %There may exist two reasons for to explain the accurary drop: 1) CLIP doesn't see many images from this domain during training time; 2) As a fashion data, there is a appearance shift between CLIP's training data set and UT-Zappo's test data set 
We suspect that CLIP may not have seen sufficient similar samples from this specific domain and therefore purely tuning the prompting is not helpful to solve the problem. In contrast, \textit{GIPCOL} adds additional compositional information to learn the element concept embedding which appears to boost the compositional learning ability within this specific domain.

%The CLIP-based models (the bottom three) already achieved satisfactory results on MIT-States dataset. But other CLIP-prompt methods, including COOP, CSP and ours, can further boost the performance, which illustrated the effectiveness of prompt learning in CZSL. However, all CLIP-based models can't achieve SOTA performance on UT-Zappos compared with CGE, one reason could be UT-Zappos is domain-specific dataset which consists of shoes and the materials. CLIP doesn't see many images from this domain during training time.

\vspace{3pt}
\noindent\textbf{Results on C-GQA.} On the more challenging C-GQA dataset, \textit{GIPCOL} also achieves new SoTA results on both closed and open world settings with an exception for the seen accuracy in the open world. However, the key metric is AUC which is consistently higher for GIPCOL in all settings.
\vspace{1pt}
\subsection{Qualitative Analysis}

\noindent \textbf{Predicted Examples}. We looked into a number of randomly selected predictions from \textit{GIPCOL} shown in Append.~\ref{quanlity_fig_append}. The red colored texts are the ground-truth labels, the blue colored texts are \textit{GIPCOL}'s correctly predicted labels and the black colored texts are \textit{GIPCOL}'s wrongly predicted labels. The first two columns present examples with correctly predicted compositional labels and the last two columns show the wrongly predicted labels, either wrong in attributes or wrong in objects. 
From this figure, we can see that \textit{GIPCOL} can recognize objects in most of the compositions in MIT-States and C-GQA datasets. However, it has difficulty to precisely predict the attributes for these two datasets. For example, it predicts \textit{modern clock} instead of \textit{ancient clock} which is the antonym of the actual attribute. But for UT-Zappos, the more domain-specific dataset, \textit{GIPCOL} even has difficulty in recognizing the objects.

%In this section, we try to explain why \textbf{GIPCOL} works in CZSL by checking the CLIP's training data. 
\noindent\textbf{Differences in Domains:} From Tables~\ref{tab:close_result} and \ref{tab:open_result}, we observe that CLIP without any prompt-tuning can achieve better performance compared to non-CLIP models on the MIT-States dataset, but not on the UT-Zappos dataset. We hypothesize that this issue can be related to the distribution difference between the pre-training data used by CLIP and the data domain of the downstream task. 
To validate this hypothesis, we further look into some concrete examples from MIT-stats and UT-Zappos. 
We take \textit{burnt boat} from MIT-Stats and \textit{Faux Fur-Shoes Clogs and Mules} from UT-Zappos for comparison as shown in Append.~\ref{clip_data_append}. The CLIP's training data is not publicly available.  However, \textit{LAION-400M}~\cite{laion400}  used the released CLIP model and obtained the closest 400M image-text pairs\footnote{https://rom1504.github.io/clip-retrieval} from their crawled dataset from Web by reverse engineering. We based our analysis on this constructed \textit{LAION-400M} subset.
By querying \textit{LAION-400M} with \textit{burn boat},  we could retrieve about $600$ relevant images.  By querying with \textit{Faux Fur\_Shoes Clogs and Mules} we can retrieve about $200$ relevant images. The first interesting difference is  in the quantity of the retrieved relevant images which is significantly lower for the shoe dataset.  The second difference is the data quality differences. As can be seen from Append.~\ref{clip_data_append}, the retrieved shoes are less similar to the UT-Zappos'  shoes when compared to the similarity of the retrieved boats to MIT-Stats boats.  
% UT-Zappos images and CLIP's pre-trained images has large discrepancy. 
We note that UT-Zappos is about shoe fashion and was constructed in 2014 while CLIP is pretrained using recent 2020's images. The change in fashion trends has made the images look different for the same compositional concept. 
Based on these observations, it is evident that the quantity and quality of  CLIP's pre-training data play an important role in its performance.

\noindent\textbf{Covering the Performance Gap}. Despite the above-mentioned issues, \textit{GIPCOL} improves the UT-Zappos dataset. While we found that CLIP's pre-training data is important in its performance in the Zero-shot setting, introducing the additional compositional knowledge in \textit{GIPCOL} positively impacts CLIP's ability in recognizing the novel compositional concepts. 
%if CLIP doesn't see many similiar examples for the target compositions, we can introduce e to help train the prompting representation and improve CLIP's compositional learning ability.
\textit{GIPCOL} uses GNN to inject compositional information into concept representations which turned out to be helpful. The improvement is important, especially for UT-Zappos which is a special domain with not many shared similar examples with CLIP's training.

\subsection{Ablation Study}

To better understand the influence of each component
in \textit{GIPCOL}, 
Tab.~\ref{component_ablation} shows the performances of its variations on UT-Zappos' closed-world setting. 
%From Table~\ref{component_ablation}, both GNN and soft-prompting are important for \textit{GIPCOL}.
%\pk{You can even remove the following sentence since you describe the details below it.}
%\pk{As it can be observed from the results, both GNN and soft-prompting impact the performance of \textit{GIPCOL}}.

\begin{table}[!htbp]
\centering
\small
\setlength\tabcolsep{3.5pt}
\begin{tabular}{l|cccc}
\toprule
%&  \multicolumn{2}{c}{Conventional ZSCL} & \multicolumn{2}{c}{Generalized ZSCL} \\
\textbf{Model} &  \textbf{S}  & \textbf{U} &  
\textbf{H}  & \textbf{AUC}\\
\toprule
\textbf{GIPCOL} & 65.0 & 68.5 & 48.8 & 36.2 \\
\hline
\textbf{- without GNN} &  64.4 & 64.0 & 46.12 & 32.2 \\
\hline
\textbf{- without prefix} & 64.7 & 62.3 & 45.9 & 31.0\\
\hline
\textbf{- without both (CLIP)} & 15.8 & 49.1 & 15.6 & 5.0 \\
\hline
\end{tabular}
\caption{Performance of \textit{GIPCOL}'s variations.}%The setting without soft-prompt and prefix vectors equals to original CLIP setting.
\label{component_ablation}
\end{table}

\noindent\textbf{Effects of GNN}. We remove the GNN module and directly set attribute and object embeddings as learnable parameters as in~\cite{prompt_comp}. The performance decreases. Especially the AUC drops from $36.2\%$ to $32.2\%$.  
%Since we need to model the interactions of element concepts to recognize the compositional concepts, GNN in \textit{GIPCOL} acts as a compositional encoder to fuse information based on the compositional graph and plays an important role in increasing the performance. 

\noindent\textbf{Effect of Learnable Prefix Vectors}. Another variant of \textit{GIPCOL} is to fix the prefix vectors and only tune the GNN module to update the class embeddings. 
%Under this variation, we still update the element concepts'  embeddings. 
From Tab.~\ref{component_ablation}, we can see that learnable prefix vectors play a more important role than GNN. In fact, adding the prefix vectors changes CLIP's textual input and makes it biased towards compositional learning, which is a key component in \textit{GIPCOL}.

%\noindent\textbf{Comparison to raw CLIP}. \jycc{what do you mean by raw? vallina CLIP?, what do you mean it has seen many compositional concepts? similar to my previous comments, if they see the concepts, cannot be described as zero shot.}Although CLIP has seen many of the compositional concepts during training, applying CLIP directly achieves no satisfactory results in CZSL. This result shows the importance of prompting learning in CZSL.

\subsection{Higher-Order Compositional Learning}

Previous work (CSP)~\cite{csp} introduced  another challenging dataset: AAO-MIT-States, a subset derived from MIT-States to evaluate the higher-order compositional learning ability in the form of attribute-attribute-object (AAO) compositions.
After learning the prefix vectors and GNN-encoded element concepts, \textit{GIPCOL} can be easily adapted to solve AAO by modifying the compositional prompt to $( \theta_1, \theta_2, ...,\theta_m, \hat{a_i}, \hat{a_j}, \hat{o_k} )$ to represent the higer-order compositions. We report the AAO results in Tab.~\ref{tab:aao}.  We can see that \textit{GIPCOL} has a better higher-order compositional leaning ability, with a $3\%$ absolute improvement compared with CSP.

\begin{table}[!htbp]
\centering
\small
\setlength\tabcolsep{3.5pt}
\begin{tabular}{ll}
\toprule
%&  \multicolumn{2}{c}{Conventional ZSCL} & \multicolumn{2}{c}{Generalized ZSCL} \\
\textbf{Model} &  Accuracy \\ 
\toprule
CLIP & 62.7\\
CSP & 72.6\\
\textit{GIPCOL} (Ours) & 75.9 \\
\bottomrule
\end{tabular}
\caption{AAO Performance of different CLIP-based models.}
\label{tab:aao}
\end{table}

\vspace*{-5mm}

\section{Conclusion}
In this paper, we propose \textit{GIPCOL}, a new CLIP-based prompting framework,  to address the compositional zero-shot learning (CZSL) problem.  The goal is to recognize compositional concepts of objects with their states and attributes as described by images.  
The objects and attributes have been observed during training in some compositions, however, the test-time compositions could be novel and unseen.
We introduce a novel prompting strategy for soft prompt construction by treating element concepts as part of a global GNN network that encodes feasible compositional information including objects, attributes and their compositions. In this way, the soft-prompt representation is influenced not only by the pre-trained VLMs but also by all the compositional representations in its neighborhood captured by the compositional graph. Our results have shown that \textit{GIPCOL} performs better and achieves SoTA AUC results on all three benchmarks including MIT-States, UT-Zappos, and C-GQA . These results demonstrate the advantages and limitations of prompting large vision and language models (such as CLIP) for compositional concept learning.   

\section*{Acknowledgement}
This project is partially supported by the Office of Naval Research (ONR) grant N00014-23-1-2417. Any opinions, findings, conclusions, or recommendations expressed in this material are those of the authors and do not necessarily reflect the views of Office of Naval Research.

%%%%%%%%% REFERENCES
{\small
\bibliographystyle{ieee_fullname}
\bibliography{egbib}
}

\clearpage
\appendix
\onecolumn
\section{Comparison between GIPCOL and CGE\label{gnn_compare}}
\begin{figure*}[ht]
\begin{center} 
\includegraphics[width=0.8\linewidth]{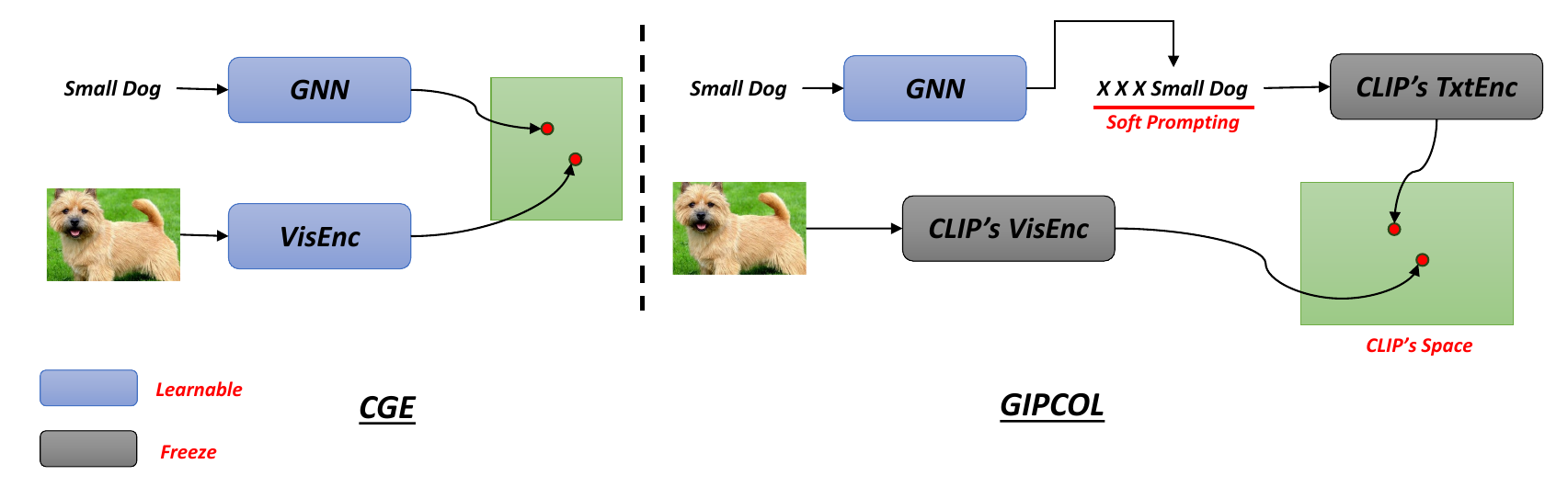}
\end{center}
\caption{Comparison between CGE and GIPCOL. GIPCOL uses GNN to help prompt construction.}
\label{cse_diff}
\end{figure*}

Although both CGE\cite{graph_comp} and \textit{GIPCOL} use GNN to encode compositional concepts, the GNN module 
functions in a fundamentally different manner in these two models. GNN in \textit{GIPCOL} helps construct the soft prompting for CZSL. However, GNN in CGE plays the text encoder role which projects the concept into the embedding space. \textit{GIPCOLL} freeze CLIP's textual and visual encoders to utilize CLIP's multi-modal aligning ability for CZSL which is more efficient. In contrast, CGE needs to train both the GNN and visual encoder to obtian competitive performance as compared in Fig.~\ref{cse_diff}.

\section{GIPCOL Algorithm\label{alg_appendix}}

\begin{algorithm}[!htbp]
\begin{algorithmic}[1]
\State Initialize GIPCOL using  CLIP's pre-trained textual and visual encoders.
\State Update element concept's representation using GNN as  Eq.~\ref{eq:gnn} and Eq.~\ref{gcn}.
\State Construct textual prompt for compositional labels using the updated element concepts and learnable prefix vectors as  Eq.~\ref{eq:prompt}.
\State Extract and normalize image/text vectors using CLIP's image/text encoder using Eq.~\ref{eq:txt_encoder} and Eq.~\ref{eq:vis_encoder} seperately.
\State Calculate the class probability as  Eq.~\ref{eq:sim} using the cosine similarity and update GIPCOL's soft-prompting layer $\Theta$ and GNN layer $\Phi$  using Cross-Entropy loss. %\jycc{what's y? $t_i$?the notation in this eq:sim has a problem. the bottom should be j from 1 to K? }
\end{algorithmic}
\caption{\textit{GIPCOL}}
\label{alg}
\end{algorithm}

\section{CZSL Dataset Statistics\label{data_stat_append}}
\begin{table}[!htbp]
\centering
\small
\setlength\tabcolsep{3.5pt}
\begin{tabular}{*{4}c}
\toprule
%&  \multicolumn{2}{c}{Conventional ZSCL} & \multicolumn{2}{c}{Generalized ZSCL} \\
&  MIT-States & UT-Zappos & C-GQA\\
\hline
\# $Attr.$ & $115$ & $16$ & $413$\\
\# $Obj.$ & $245$ & $12$ & $674$\\
\# $Attr. \times Obj.$ & $28175$ & $192$ & $278362$\\
\hline
\# Train Pair & $1262$ & $83$ & $5592$\\
\# Train Img. & $30338$ & $22998$ & $26920$\\
\hline
\# Val. Seen Pair & $300$ & $15$ & $1252$\\
\# Val. Unseen Pair & $300$ & $15$ & $1040$\\
\# Val. Img. & $10420$ & $3214$ & $7280$\\
\hline
\# Test Seen Pair & $400$  & $18$ & $888$\\
\# Test Unseen Pair & $400$  & $18$ & $923$\\
\# Test Img. & $19191$ & $2914$ & $5098$\\
\hline
\end{tabular}
\caption{Dataset Statistics for MIT-States, UT-Zappos and C-GQA.}
\label{data_split}
\end{table}

\section{Feasible Score Threshold in Open-World CZSL\label{feas_score_append}}

In open-world CZSL (OW-CZSL), we use the validation set to choose a feasible threshold to remove less feasible compositions from the output space and the adopted threshold in \textit{GIPCOL} is shown in Tab.~\ref{feas_score}.

\begin{table}[!htbp]
\centering
\begin{tabular}{*{2}c}
\toprule
%&  \multicolumn{2}{c}{Conventional ZSCL} & \multicolumn{2}{c}{Generalized ZSCL} \\
Dataset &  Feasibility Score\\
\hline
MIT-States & 0.40691\\
UT-Zappos & 0.51878\\
C-GQA & 0.49941\\
\hline
\end{tabular}
\caption{\textit{GIPCOL}'s feasibility threshold score.}
\label{feas_score}
\end{table}

% \clearpage
\section{Qualitative Examples\label{quanlity_fig_append}}
\begin{figure*}[ht]
\begin{center} 
\includegraphics[width=0.9\linewidth]{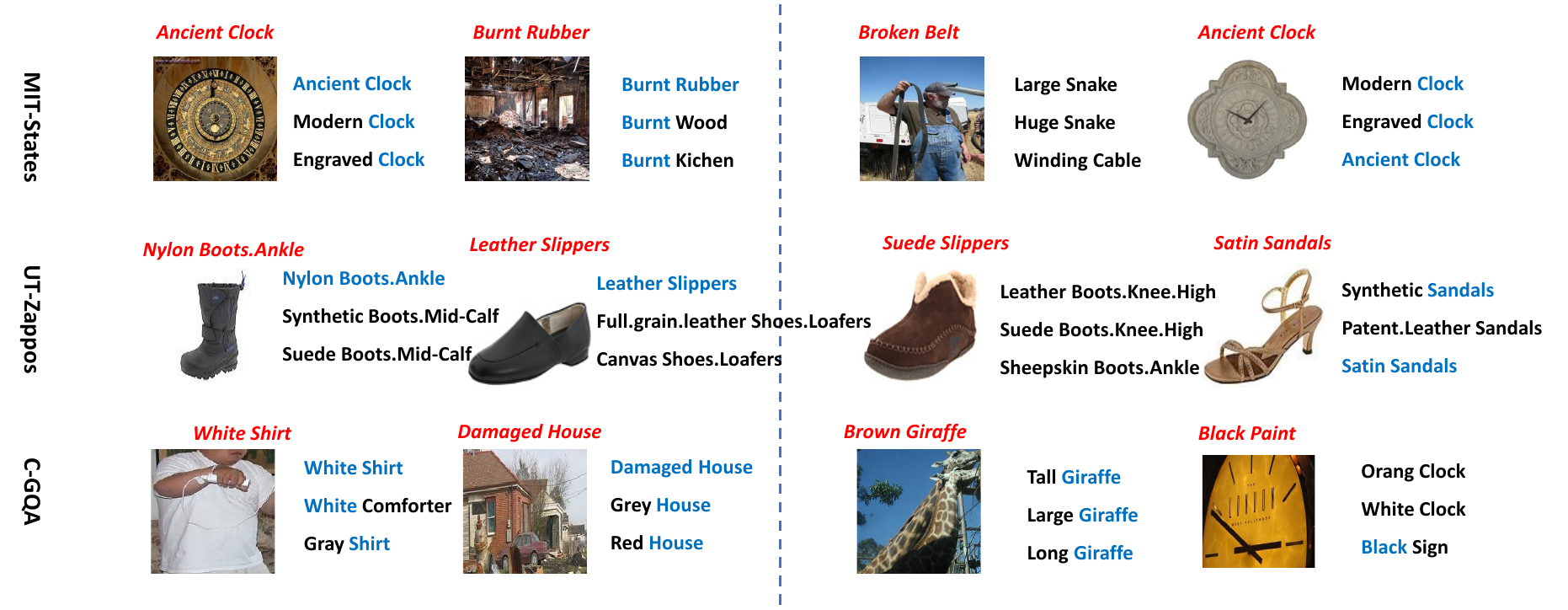}
\end{center}
\caption{We show the top-3 predictions of our proposed model for some images. Red colors are ground-truth labels, blue colors are correctly predicted labels and black colors are wrongly predicted labels. }
\label{quanlity_fig}
\end{figure*}

\section{Comparison between CLIP's Pre-train Dataset and Target Dataset\label{clip_data_append}}

\begin{figure*}[ht]
\begin{center} 
\includegraphics[width=0.8\linewidth]{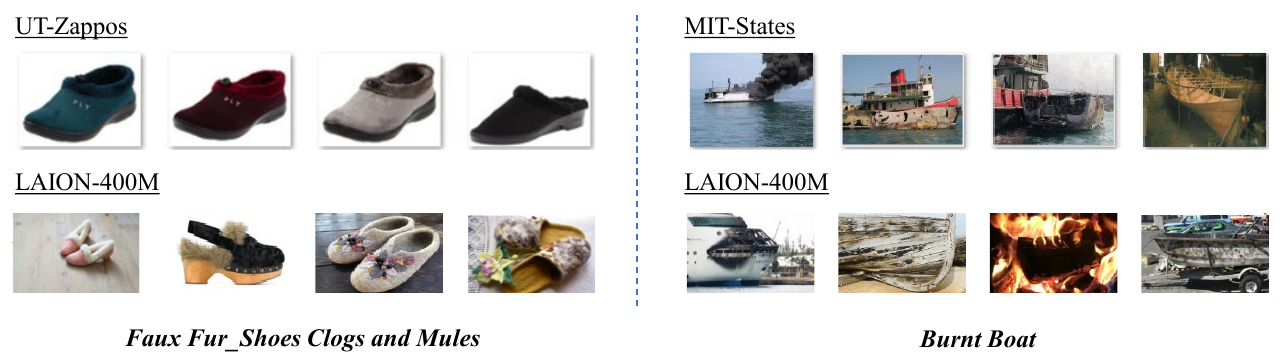}
\end{center}
\caption{Comparison between retrieved images from Laion400M and UT-Zappos/MIT-States.}
\label{clip_data}
\end{figure*}

We visualize CLIP's pre-training dataset and target domain dataset in Fig.~\ref{clip_data}.
From this figure, we can see that MIT-States have similar visual appearance with CLIP's pre-trained data. However, for UT-Zappos, because of the fashion style change overtime, shoes have significant visual appearance between the pre-training dataset and the target dataset. Results in Tab.~\ref{tab:close_result} and Tab.~\ref{tab:open_result} have shown the domain similarity plays an important role in prompting-based method. Prompting CLIP without any training can achieve better performance on MIT-State then UT-Zappos.
\textit{GIPCOL} helps address this challenge partially by prompting design based on the restuls.

\clearpage

\end{document}